\documentclass[runningheads]{llncs}
\usepackage[dvipsnames]{xcolor}
\usepackage{tikz}
\usepackage{comment}
\usepackage{amsmath,amssymb} 
\usepackage{booktabs}
\usepackage{cases}
\usepackage[accsupp]{axessibility}  
\usepackage{graphicx}
\usepackage{float}

\newcommand{\drop}[1]{\textcolor{red}{\scriptsize{($\downarrow$#1)}}}
\newcommand{\rise}[1]{\textcolor{blue}{\scriptsize{($\uparrow$#1)}}}

\newcommand{\Rise}[1]{\textcolor{blue}{\scriptsize{(\bf $\uparrow$#1)}}}

\newfloat{figtab}{htb}{fgtb}
\makeatletter
  \newcommand\tabcaption{\def\@captype{table}\caption}
  \newcommand\figcaption{\def\@captype{figure}\caption}
\makeatother

\usepackage[pagebackref=true,breaklinks=true,letterpaper=true,colorlinks,bookmarks=false, citecolor=ForestGreen]{hyperref}

\usepackage{orcidlink}

\begin{document}
\pagestyle{headings}
\mainmatter
\def\ECCVSubNumber{3139}  

\newcommand\blfootnote[1]{%
  \begingroup
  \renewcommand\thefootnote{}\footnote{#1}%
  \addtocounter{footnote}{-1}%
  \endgroup
}

\title{Semi-Supervised Single-View 3D Reconstruction via Prototype Shape Priors} 

\titlerunning{Semi-Supervised Single-View 3D Reconstruction}

\author{Zhen Xing$^{1,2}$\orcidlink{0000-0001-6407-0321} \and Hengduo Li$^{3}$\orcidlink{0000-0001-5314-6853} \and  Zuxuan Wu$^{1,2\dagger}$\orcidlink{0000-0002-8689-5807} \and Yu-Gang Jiang$^{1,2}$\orcidlink{0000-0002-1907-8567}
}

\authorrunning{Zhen Xing et al.}
\institute{$^{1}$~Shanghai Key Lab of Intell. Info. Processing, School of CS, Fudan University \\
$^{2}$~Shanghai Collaborative Innovation Center on Intelligent Visual Computing
$^{3}$~University of Maryland\\
}
\maketitle

\begin{abstract}
\blfootnote{$^{\dagger}$ Corresponding author.}
The performance of existing single-view 3D reconstruction methods heavily relies on large-scale 3D annotations. However, such annotations are tedious and expensive to collect. Semi-supervised learning serves as an alternative way to mitigate the need for manual labels, but remains unexplored in 3D reconstruction. Inspired by the recent success of  semi-supervised image classification tasks, we propose SSP3D, a semi-supervised framework for 3D reconstruction. In particular, we introduce an attention-guided prototype shape prior module for guiding realistic object reconstruction. We further introduce a discriminator-guided module to incentivize better shape generation, as well as a regularizer to tolerate noisy training samples. On the ShapeNet benchmark, the proposed approach outperforms previous supervised methods by clear margins under various labeling ratios, (\emph{i.e.}, 1\%, 5\% , 10\% and 20\%). Moreover, our approach also performs well when transferring to real-world Pix3D datasets under labeling ratios of 10\%.  We also demonstrate our method could transfer to novel categories with few novel supervised data. 
Experiments on the popular ShapeNet dataset show that our method outperforms the zero-shot baseline by over 12\% and we also perform rigorous ablations and analysis to validate our approach.  Code is available at \href{https://github.com/ChenHsing/SSP3D}{https://github.com/ChenHsing/SSP3D}.

\end{abstract}

\section{Introduction}

Reconstructing 3D shape from RGB images plays an important role in many applications, such as 3D printing, virtual reality and 3D scene understanding. Human can easily infer 3D shape and scene object from single-view images mainly because of the powerful shape priors of human visual systems, yet it remains challenging to model such strong priors for accurate single-view 3D reconstruction.
While Structure From Motion(SFM)~\cite{sfm} and Simultaneous Localization and Mapping (SLAM)~\cite{slam} are feasible solutions, they require abundant data annotations and inferring camera parameters.

Recently, with the growing interest in deep learning, great success has been achieved in predicting 3D shape from a single image with deep Convolutional Neural Networks (CNNs) \cite{3dr2n2,pixel2mesh,pix2vox}. But there are still limitations of these methods: (i) The astounding performance comes at the cost of massive amount of labeled images with fine-grained 3D shape, which is time-consuming and labour-intensive to obtain. (ii) Inferring 3D shape from a single image is an ill-posed problem because there are multiple plausible shapes given a 2D image.

\begin{figure}[ht]
\centering
\includegraphics[width=1.0\columnwidth]{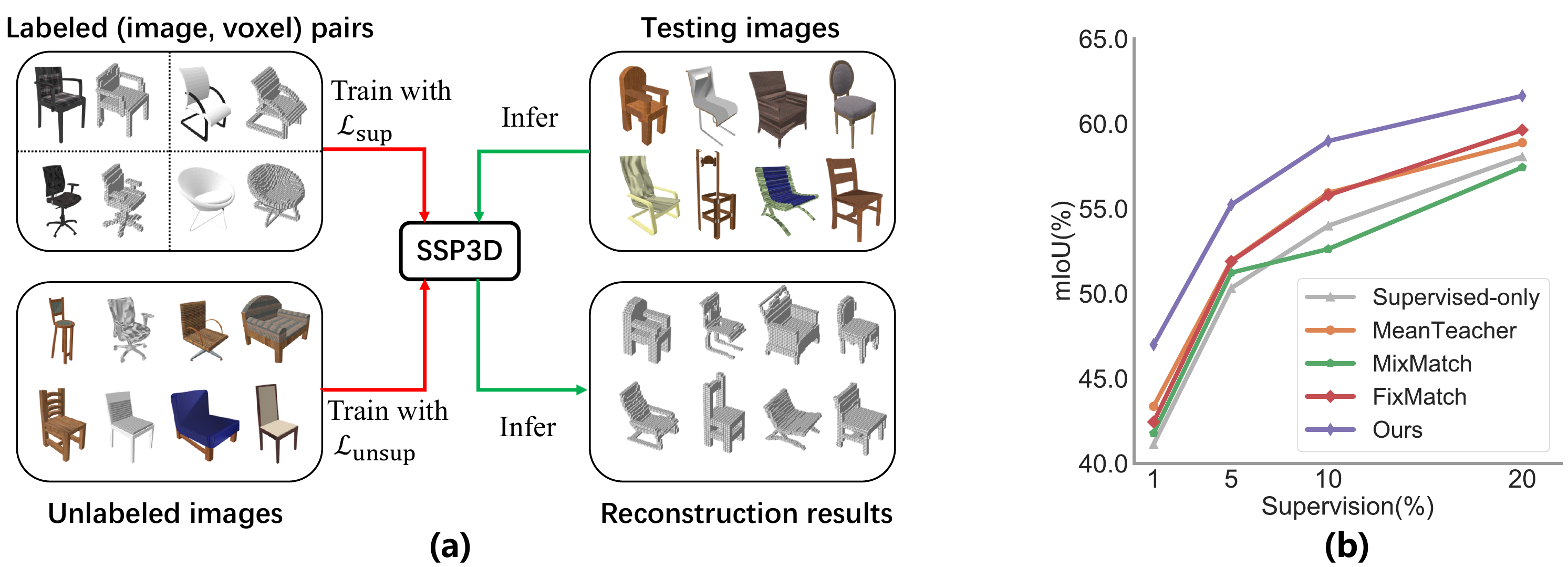} 
\caption{ (a) Illustration of semi-supervised single-view 3D reconstruction. Our SSP3D can predict 3D shape for an unlabeled image after training with a mixture of labeled data and unlabeled data. (b) Our proposed model can efficient leverage the unlabeled data and outperform supervised-only method and state-of-the-art semi-supervised image recognition extended methods.}
\label{fig01}
\end{figure}

Semi-Supervised learning (SSL) is a popular strategy to learn in the low-data regime by leveraging the readily available unlabeled data, which has demonstrated great success for image classification~\cite{meanteacher,mixmatch,weng2021semi} and object detection~\cite{liu2021unbiased}. Generalizing best practices~\cite{meanteacher,sohn2020fixmatch} that work well in the 2D domain to 3D reconstruction, while appealing, is challenging. On one hand, it remains unclear how to evaluate the quality of 3D shape pseudo labels, which are the core for SSL. On the other hand, inferring the actual 3D shape of an object from a single image requires strong shape priors, yet existing single-view 3D reconstruction methods~\cite{3dr2n2,pix2vox} require a large amount of annotated data to learn the shape priors implicitly with the model parameters. As a result, the 3D reconstruction network trained with limited annotations will likely produce low-quality reconstruction results, especially for the images with heavy occlusion.

To tackle these challenges, we propose a semi-supervised learning framework with several components specially designed for single-view 3D reconstruction as shown in Figure \ref{fig01}. Inspired by the recent advances in SSL for image classification~\cite{meanteacher,sohn2020fixmatch}, we use the teacher-student pseudo labeling method as the training paradigm of our framework. In order to generate more reliable pseudo labels for the unlabeled images, we use a Prototype Attentive Module for providing shape priors explicitly. In particular, we first obtain 3D prototype shape as candidate shape priors through clustering algorithms (\emph{e.g.}, KMeans). For a given image, we extract the image feature through a 2D encoder. The relationship of image feature and 3D prototype is captured with the help of the attention mechanism to obtain the shape priors, which serve as a bridge to encourage perceptually realistic reconstruction and prevent mode collapses~\cite{3dr2n2,pix2vox}. 

In addition, we introduce a module named Shape Naturalness Module that serves as a discriminator distinguishing predicted 3D shapes from ground-truth 3D shapes. During training, an additional loss is used to penalize unnatural reconstruction results from the model in a generative adversarial learning manner such that the model is incentivized to generate more realistic 3D shapes. Meanwhile, the output of the discriminator can be directly used as an approximation of the quality of pseudo labels such that the inaccurate pseudo labels can be ignored or down-weighted accordingly when training the student model.

\noindent In conclusion, the main contributions of this paper is summarized as follows:
\begin{itemize}

\item We propose a semi-supervised prototype 3D reconstruction network (SSP3D) to reconstruct 3D shapes from a single RGB image. Our work is the first attempt to reconstruct 3D volume in semi-supervised learning with only $1\%$ labeled data of train set.

\item Without additional information, an effective yet lightweight shape prior fusion module is proposed, which can be easily incorporated into 3D reconstruction networks with similar architecture. In addition, the discriminator module we proposed guides the generation of natural shapes and serves as a scorer to filter out noisy training samples for the student model.

\item We are the first to establish a semi-supervised benchmark to measure the single-view 3D reconstruction network. Experiments show that our model achieves the state-of-the-art on two datasets and settings under various  labeling ratios. 
We hope that our results serve a strong baseline to encourage future research in more robust semi-supervised 3D reconstruction methods.
\end{itemize}

\section{Related Work}
\noindent\textbf{Deep Learning for 3D Reconstruction}
Recently,  deep learning techniques have been widely used for 3D  reconstruction. 3D-R2N2~\cite{3dr2n2} is among the earliest work exploring the 3D reconstruction based on Recurrent Neural Network. It establishes a benchmark for 3D reconstruction with a synthetic ShapeNet dataset.  3D-VAE-GAN~\cite{vae-gan} builds upon Variational Autoencoders (VAE) and Generative Adversarial Networks (GANs) to reconstruct 3D shapes. OGN~\cite{octree} and Matryoshka Networks~\cite{matryoshka} use octree and nested shape layers to represent 3D volumes of objects, respectively. Marrnet~\cite{marrnet}, ShapeHD~\cite{shapehd} and GenRe~\cite{genre} adopt 2.5D information such as depth, silhouette and surface normal of RGBs as intermediate shape priors to reconstruct 3D shapes. Pix2Vox~\cite{pix2vox} and Pix2Vox++~\cite{pix2vox++} build robust backbones for 3D volume reconstruction and achieve state-of-the-art results with encoder-decoder architectures. Mem3D~\cite{mem3d} requires a great extra storage space to provide shape priors, which limits its applicability.  EVolT~\cite{evolt} and 3D-RETR~\cite{3d-retr} leverage transformers as backbone networks to reconstruct 3D shapes. Unlike most existing work that are trained in a supervised manner, we explore semi-supervised learning for 3D reconstruction.

\noindent\textbf{Deep Semi-Supervised Learning}
The overall purpose of semi-supervised learning (SSL) is to effectively use unlabeled data without relying on any manual supervision to expand supervised learning when the labeled training data is scarce. Recent semi-supervised methods mainly contain two principles: data augmentations and consistency regularization. The model is expected to be consistent and robust to data augmentations---producing consistent outputs for the original and augmented inputs. Many methods use different data augmentations~\cite{mixmatch,temporal,regularization} or dropout~\cite{meanteacher} of models to generate images of different transformations. Researchers also use multiple networks to generate different views of the same input data~\cite{cotraining}, or mix input data to generate training data and labels~\cite{mixup,cutmix,guo2019mixup,augmix}. In single-view 3D reconstruction, Semi-supervised Soft Rasterizer (SSR)~\cite{ssrpose} and~\cite{limitedpose} try to reconstruct 3D objects with few amount of annotation data, but they all rely on the annotations of additional camera pose or silhouette. To the best of knowledge, the settings of SSL with only single-view image have not been studied in 3D reconstruction, a complex and challenging task that depends on fine-grained human annotations.

\section{Method}
\begin{figure}[ht]
\centering
\includegraphics[width=0.82\columnwidth]{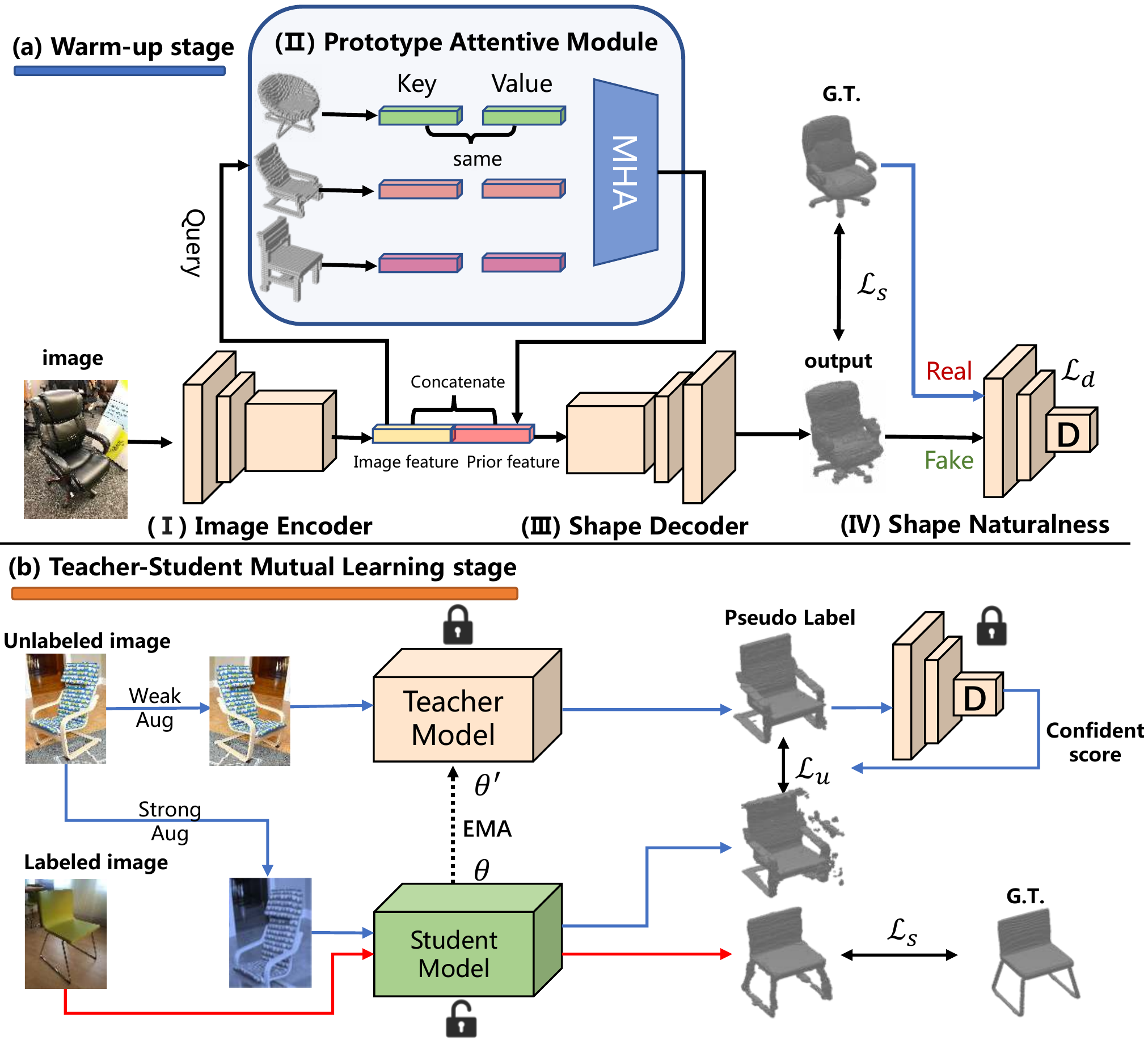} 
\caption{\textbf{ Overview of Our SSP3D.} SSP3D consists of two stages.  \textbf{Warm-up}: we use available supervised data to train 3D reconstruction network. \textbf{Teacher-Student mutual learning stage}: for unsupervised data, \emph{Teacher} with fixed parameters generate pseudo-labels to train \emph{Student}. At the same time, \emph{Teacher} and \emph{Student} are given weakly and strongly augmented inputs respectively. In order to avoid the interference of pseudo-labels noise generated by the \emph{Teacher}, we give a confidence score weight to the unsupervised loss by discriminator. The knowledge learned by \emph{Student} online is slowly transferred to the weight replication mode of \emph{Teacher} through exponential moving average (EMA). When the reconstruction network is trained and converged in the Warm-up stage, we switch to the Teacher-Student mutual learning stage.
}
\label{fig02}
\end{figure}

\noindent\textbf{Problem Definition}
For a single-view image $x$ of any object, the goal is to reconstruct the 3D shape $y$ of the object. As discussed earlier, current methods for single-view 3D reconstruction typically require large amount of annotations that are time-consuming and labour-intensive to obtain. We thus explore developing a semi-supervised learning framework for the task to alleviate the need of annotated data during training. 

Suppose we have $N$ training samples, including $N_L$ labeled image-3D pairs $(x_l, y_l) \in D_ L$ and $N_ U$ unlabeled image data $(x_u) \in D_U$. As in prior work, $D_ L$ and $D_ U$ are sampled from the same data distribution (\emph{e.g.}, either synthetic or real-world). Our purpose is to leverage $D_L$ and $D_U$ together to train the model for an improved performance on reconstructing the 3D shapes of objects. 

\noindent\textbf{Overview} As shown in Figure~\ref{fig02}, our framework SSP3D contains two training stages: Warm-up stage and Teacher-student mutual learning stage. In the Warm-up stage, the available labeled set $D_L$ is used to train a ``teacher'' model; in the Teacher-student mutual learning stage, the teacher model first generates pseudo labels (\emph{i.e.}, predicted 3D shapes) for the unlabeled set $D_U$, and then a ``student'' model -- initialized from the pre-trained teacher model -- is trained on $D_L$ and $D_U$ for an improved performance. For effective distillation, strong data augmentation is applied on the input to student model. The teacher model also temporally aggregates the weights from the student model to produce more refined pseudo labels. 

While appealing, directly extending existing SSL methods like MeanTeacher~\cite{meanteacher} and FixMatch~\cite{sohn2020fixmatch} for single-view 3D reconstruction is challenging since the pseudo labels from the teacher model can be quite noisy for two main reasons: 1) inferring accurate 3D shape from single-view image requires strong prior that is difficult to learn without massive annotated data; 2) it is unclear how to evaluate the quality of the predicted pseudo 3D shapes to filter out inaccurate predictions. To this end, we propose two modules namely Prototype Attention Module and Shape Naturalness Module to address these challenges.

In the following text, we first introduce our proposed model components in Warm-up stage (Sec.~\ref{section:warmup}), and then we show how the Teacher-student mutual learning stage works with pseudo labelling and teacher refinement methods (Sec.~\ref{section:teacherstudent}). Finally, we elaborate the optimization of our framework in Sec.~\ref{section:trainingparadigm}.

\subsection{Warm-up stage}\label{section:warmup}
As shown in Fig.~\ref{fig02}, SSP3D consists of four modules, among which image encoder and shape decoder are consistent with the state-of-the-art method Pix2Vox~\cite{pix2vox}, whereas the proposed prototype attentive module and shape naturalness module will be presented below. At this stage, the teacher model is trained on labeled set $D_L$ in standard supervised learning manner. 

\begin{figure}[ht]
\centering
\includegraphics[width=0.8\columnwidth]{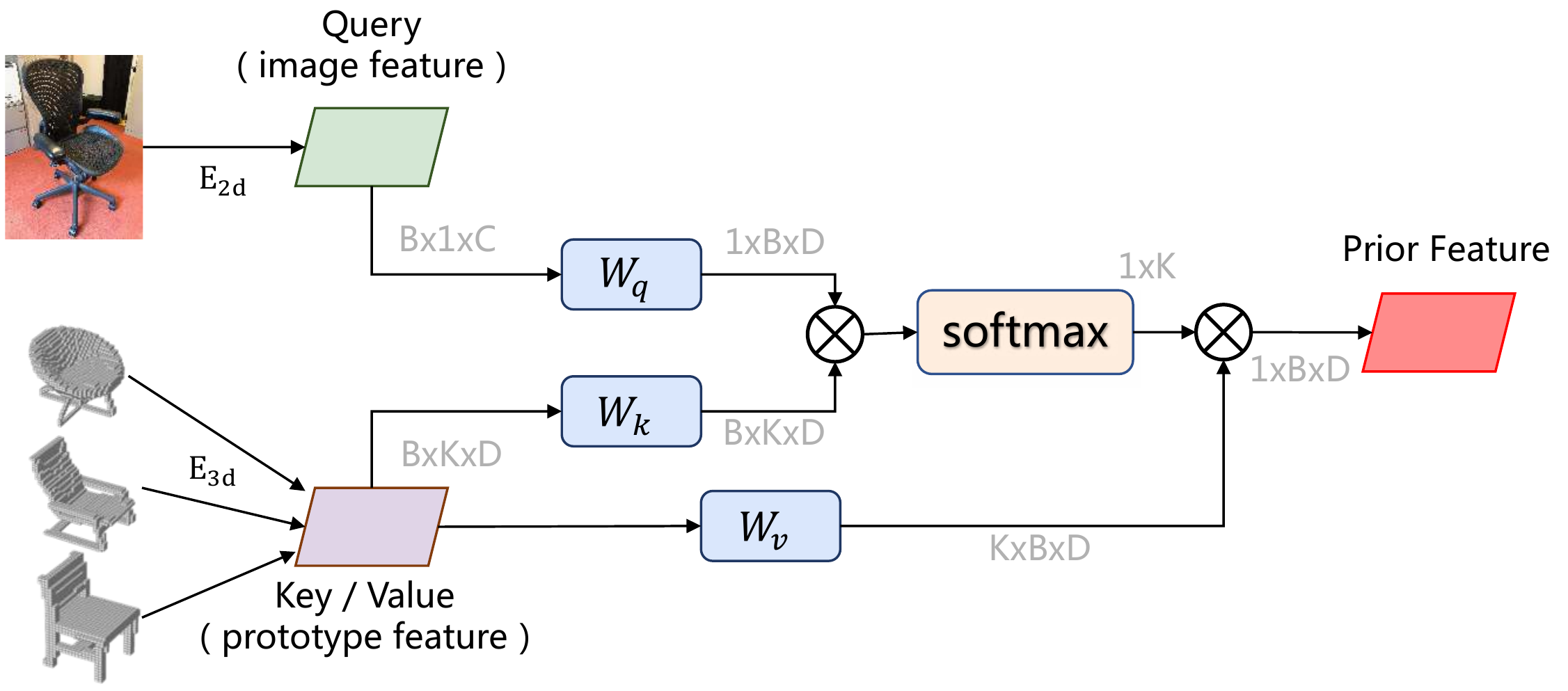} 
\caption{\textbf{ Overview of Our Prototype Attentive Module.} 
}
\label{fig03}
\end{figure}

\noindent\textbf{Prototype Attentive Module} In most existing work on single-view 3D reconstruction, the shape priors are learned implicitly with the parameters of the model~\cite{3dr2n2,pix2vox,evolt}, which may lead to poor performance for some noisy or occluded images~\cite{3dr2n2}, especially when annotated training data is not abundant. Therefore, standard 3D reconstruction models are likely to produce noisy and inaccurate pseudo labels when used as the teacher model directly, resulting in poor performance when training the student model under such semi-supervised learning setting. 

To tackle this problem, we propose to augment the image features with learned category-specific shape priors explicitly, so that the strong priors learned from labeled data could help infer more realistic and natural object shapes.

For supervised data, we obtain the shape prototype ${P_i}^k $ of the specified categories by doing K-Means clustering on the features learned by a 3D Autoencoder~\footnote{Please refer to Appendix for more details.}, and we take the clustering center of these categories as the prototype shape priors. The designed attention-based shape priors acquisition mechanism is shown in Fig.~\ref{fig03}. Firstly, image encoder extracts the 2D feature of the image as query in Eq.~\ref{e2d}. Secondly, we extract feature of the prototype through 3D encoder to obtain the prototype feature in Eq.~\ref{e3d}, which is used as the key and value in the attention mechanism~\cite{vaswani2017attention}. We then use three separate linear layers parameterized by $W_q$, $W_k$ and $W_v$ to extract query, key, value embedding Q, K and V in Eq.~\ref{transforms}. Formally, the shape prior feature can be obtained by multi-head attention (MHA)~\cite{vaswani2017attention} in Eq.~\ref{MHA}.

\begin{equation}
\label{e2d}
 \text{Image features:~} Query = \texttt{Encoder2d}(I_q),
\end{equation}
\begin{equation}
\label{e3d}
  \text{Prototype features:~} Key, Value = \texttt{Encoder3d}(P_i),
\end{equation}
\begin{equation}
\label{transforms}
 Q = Query\cdot W_q, ~~  K = Key\cdot W_k, ~~   V = Value\cdot W_v,
\end{equation}
\begin{equation}
\label{MHA}
    \text{Prior features} = \texttt{MHA} (Q, K, V ).
\end{equation}

Here, $I_q$ is the query image, $P_i$ indicates the prototype 3D voxel, $W_q$ $\in \mathbb{R}^{C \times D}$, $W_k$, $W_v$ $\in \mathbb{R}^{D \times D}$ are learnable matrices. In the previous work using shape priors~\cite{wallace2019few,mem3d}, 3D voxel can be directly used as shape priors in the form of additional inputs, however they can not capture the correlations between the images and multiple prototype shape priors. In contrast, we use the attention-based module to extract the shape priors by exploring the association between image features and 3D prototypes.

\noindent\textbf{Shape Naturalness Module}
The shape reconstruction network typically uses only one supervised loss during training, yet the inherent uncertainty of the loss will lead to unrealistic and inaccurate prediction of object shapes especially on object surface. 

Inspired by~\cite{shapehd}, we develop a shape naturalness module that servers as a discriminator distinguishing predicted shape and the corresponding ground-truth shape, and penalizes the network in an adversarial learning manner when unnatural shapes are generated. 

Unlike~\cite{vae-gan} and~\cite{shapehd} which use a pre-trained 3D-GAN as a discriminator to judge whether a shape is real, our framework is learned in an end-to-end generative adversarial training manner. In particular, we take parts (I)-(III) in Fig.~\ref{fig02} as the generator, and the shape naturalness module is used to distinguish the generated shape from the real shape. The optimization is achieved by minimizing the following loss $\mathcal{L}_d$: 

\begin{equation}
 {\cal{L}}_d = {\mathbb{E}}_{y_{p} \sim D_{p}} \rm{log}  \emph{D}(y_{p}) + \mathbb{E}_{y_{g}\sim D_{g}} \rm{log} (1-\emph{D}(y_{g})),
\end{equation}
where $D_p$ and $D_g$ are predicted and groundtruth distributions, $y_p$ and $y_g$ are samples in $D_p$ and $D_g$ respectively, and $D$ is the discriminator here.

\subsection{Teacher-Student Mutual Learning stage}\label{section:teacherstudent}

\noindent\textbf{Overview} After the teacher model converges in the Warm-up stage, it is used to produce pseudo labels on unlabeled images to supervise the student model. For effective and efficient distillation, we initialize the student model with the weights of the teacher model and apply strong data augmentations on input images to the student following the common SSL paradigm~\cite{sohn2020fixmatch}. On the other hand, the teacher takes weakly augmented images as inputs and aggregates the weights of the student temporally throughout the Teacher-Student Mutual Learning stage to generate more reliable pseudo labels.

\noindent\textbf{Student Learning} To utilize the readily available unlabeled images $D_u$, we use the pseudo-labeling method to generate labels for $D_u$ to train the student model, which has been shown effective for semi-supervised image classification~\cite{meanteacher,sohn2020fixmatch} and object detection~\cite{liu2021unbiased,rethinkingsemiod}. 

Formally, for unsupervised data, the teacher model first generates the soft label $\hat{I}$ in voxel, in which each voxel entry belongs to $[0,1]$. We first binarize it into hard labels, where each entry in the 3D voxel is binarized as follows:
\begin{numcases}
{I(i,j,k)=}
    1, &  $\hat{I}(i,j,k) > \delta$\\
    0, &  otherwise
\end{numcases}

We then train the student model by taking the binarized pseudo label $I$ as the ground truth. In addition, we jointly train students with the same amount of unsupervised and supervised data in each mini-batch to ensure that the model is not biased by pseudo labels.

\noindent\textbf{Confidence Scores for Pseudo Label}
The predictions from the teacher model are more or less inaccurate compared with the ground-truth shapes. Therefore, a filtering mechanism is desired to keep only the mostly accurate predictions as pseudo labels to train the student model. Existing semi-supervised classification methods often use the confidence scores predicted by the network as a proxy and only keep the confident predictions as pseudo labels through applying pre-defined thresholds~\cite{yalniz2019billion} or using Top-k selection~\cite{sohn2020fixmatch}. However, such confidence scores are missing in 3D reconstruction, and a new solution is needed to measure the quality of the generated 3D pseudo labels.

To this end, the shape naturalness module is also designed to serve as a naturalness ``scorer'' directly. In particular, the sigmoid-normalized output of the discriminator naturally indicates the possibility that an output sample is real or fake since the discriminator is optimized by a binary cross entropy loss using label $1$ for real ground truth, and $0$ for generated shape. We therefore use this output as the confidence score to measure the quality of generated pseudo label. The confidence score can be used to reweight the unsupervised loss, which will be described in detail in Section \ref{section:trainingparadigm}.

\noindent\textbf{Teacher Refinement}
In order to obtain more refined pseudo labels, we use exponential moving average (EMA) to gradually update the teacher model with the weights of the student model. The slow updating process of teacher model can be considered as an ensemble of student models at different training time stamps. The update rule is defined below:
\begin{equation}
\label{ema}
{\theta}_t \gets \alpha {\theta}_t +(1- \alpha) {\theta}_s,
\end{equation}
where $\alpha$ is momentum coefficient. In order to make the training process more stable, we slowly increase $\alpha$ to 1 through cosine design as in~\cite{cosinema}. This method has been proved to be effective in many existing works, such as self-supervised learning~\cite{he2020momentum,grill2020bootstrap}, SSL image classification~\cite{meanteacher} and SSL object detection method~\cite{liu2021unbiased,rethinkingsemiod}. Here, we are the first to introduce it and validate its effectiveness in semi-supervised 3D reconstruction to the best of our knowledge.

\subsection{Training paradigm}\label{section:trainingparadigm}

The training process is completed in two stages. In the Warm-up stage, we adopt reconstruction loss and GAN loss jointly and train the teacher model on $D_L$. In the Teacher-Student mutual learning stage, the generator part is duplicated as two models (Teacher and Student). The parameters of teacher and discriminator are fixed in this stage. We only optimize students through supervised and unsupervised losses. 

\noindent\textbf{Reconstruction Loss}
For the 3D reconstructions network, both the reconstruction prediction and the ground truth are in the form of voxels. We follow previous works~\cite{wallace2019few,cgce,pix2vox,pix2vox++} that adopt binary cross entropy loss as the reconstruction loss function:
\begin{equation}
\label{lossrec}
{\cal{L}}_{rec}=\frac{1}{r_{v}^{3}}\sum_{i=1}^{r_{v}^{3}}[{gt}_i\log({pr}_i)+(1-{gt}_i)\log(1-{pr}_i)],
\end{equation}
where $r_v$ represents the resolution of the voxel space, $pr$ and $gt$ represent the predict and the ground truth volume.

\noindent\textbf{Warm-up Loss} In the Warm-up stage, all parts of the models are end-to-end trained on labeled set $D_L$. The objective function is:
\begin{equation}
   \underset{\theta_f}{\text{min}}\,\, \underset{\theta_d}{\text{max}}\,\, {\cal{L}}_{rec}(\theta_f) + \lambda_d {\cal{L}}_{d}(\theta_d).
\end{equation}
Where $\theta_f$ and $\theta_d$ are the parameter of generator and discriminator, respectively. $\lambda_d$ is the balance parameter of loss terms. We set $\lambda_d$ to 1e-3 here. 

\noindent\textbf{Teacher-Student Mutual Loss}
At the second stage, for supervised data, we use the BCE loss function as in Eq.~\ref{lossrec}. For unlabeled data, we use the loss function below:
\begin{equation}
    {\cal{L}}_{unsup}=  \sum_{i=1}^{n}{\text{score}}_i(\hat{y_i} - y_i)^2,
\end{equation}
where $y_i$ and $\hat{y_i}$ are the target and predicted shapes, respectively. ${score}_i$ denotes the confidence score of $\hat{y_i}$ output by the discriminator. Note that we used squared L2 loss or the Birer score~\cite{brier1950verification} instead of binary cross entry loss in the optimization of unsupervised data. The Brier score is widely used in semi-supervised literature because it is bounded and does not severely penalize the probability of being far away from the ground truth. Our initial experiments show that square L2 loss results in slightly better performance than binary cross entropy.

The loss function for training the student model is shown below:
\begin{equation}
 {\cal{L}}={\cal{L}}_{rec}+ \lambda_u {\cal{L}}_{unsup}.
\end{equation}
where $\lambda_u$ is the balance parameter of loss terms, which is set as $5$ here. Through the joint training of supervised loss and unsupervised loss, we can make full use of labeled and unlabeled data to achieve better performance.

\section{Experiments}

\subsection{Experimental Setup}
\noindent\textbf{Datasets} We use ShapeNet~\cite{chang2015shapenet} and Pix3D~\cite{sun2018pix3d} in our experiments. The ShapeNet~\cite{chang2015shapenet} is described in 3D-R2N2~\cite{3dr2n2}, which has 13 categories and 43,783 3D models. Following the split defined in Pix2Vox~\cite{pix2vox}, we randomly divide the training set into supervised data and unlabeled data based on the ratio of labeled samples, \emph{i.e.}, 1\%, 5\%, 10\% and 20\%. The voxel resolution of ShapeNet is $32^3$. Pix3D~\cite{sun2018pix3d} is a large-scale benchmark with image-shape pairs and pixel level 2D-3D alignment containing 9 categories. We follow the standard S1-split, which contains 7,539 train images and 2,530 test images as in Mesh R-CNN~\cite{meshrcnn}. Because Pix3D is loosely annotated (\emph{i.e.}, an image may contain more than one object but only one object is labeled), we use ground-truth bounding boxes to cut all the images as \cite{pix2vox,pix2vox++}. Similarly, we randomly sample 10\% of the training set as labeled data and use the remaining samples as unlabeled data. The voxel resolution of Pix3D is $128^3$ and we have also changed the network parameters accordingly following common practice~\cite{pix2vox++}.

\noindent\textbf{Evaluation Metric}
We used Intersection over Union (IoU) for the evaluation metric as in ~\cite{3dr2n2,pix2vox}. It is defined as follows:
\begin{equation}
\text{IoU}=\frac{\sum_{i,j,k}{\cal{F}}(\hat{p}_{(i,j,k)}>t){\cal{F}}(p_{(i,j,k)})}{\sum_{i,j,k}{\cal{F}}[{\cal{F}}(\hat{p}_{(i,j,k)}>t)+{\cal{F}}(p_{(i,j,k)})]},
\end{equation}
where $\hat{p}_{(i,j,k)} $ and $p_{(i,j,k)}$ represent the predicted possibility and the value of ground truth at voxel entry $(i,j,k)$, respectively. $\cal{F}$ is a shifted unit step function and $t$ represents the threshold, which is set to 0.3 in our experiments.

\noindent\textbf{Implementation details}
In both stages, the batch size is set to 32, and the learning rate decays from $1e-3$ to $1e-4$. We use Adam~\cite{kingma2014adam} as the optimizer. We set $\alpha$ to 0.9996, the number of clusters for prototypes to 3, and the number of multi-head of attention to 2. The $\delta$ is set to 0.3. We train the network for 250 epochs in the Warm-up stage and 100 epochs in the Teacher-Student mutual learning stage.

\subsection{Main Results}
\noindent\textbf{Baseline}
We compare our approach with various baselines and direct extensions of popular semi-supervised approaches for 2D image classification. Firstly, we consider the encoder-decoder architecture of Pix2Vox~\cite{pix2vox} as our supervised baseline. Note that we change the backbone from VGG19~\cite{vgg} to ResNet-50~\cite{he2016deep} for decreasing parameters following Pix2Vox++~\cite{pix2vox++}. Secondly, we extend state-of-the-art SSL methods for image classification such as MeanTeacher~\cite{meanteacher}, MixMatch~\cite{mixmatch} and FixMatch~\cite{sohn2020fixmatch}, to the task of 3D reconstruction, which serve as strong semi-supervised baselines. We use the same backbone and experimental settings for all the
baselines and our approach for fair comparisons. More details of implementation could be found in Appendix.

\begin{table}[]
\caption{\textbf{Comparisons of single-view 3D object reconstruction on ShapeNet at $32^3$ resolution with different labeling ratios.} We report the mean IoU $(\%)$ of all categories. The best number for each category is highlighted in bold.}
\centering
\scalebox{0.9}{

\begin{tabular}{lclclclcl}
\midrule
Approach\textbackslash split & \multicolumn{2}{c}{\begin{tabular}[c]{@{}c@{}}1\%\\ 301 labels\\ 30596 images\end{tabular}} & \multicolumn{2}{c}{\begin{tabular}[c]{@{}c@{}}5\%\\ 1527 labels\\ 30596 images\end{tabular}} & \multicolumn{2}{c}{\begin{tabular}[c]{@{}c@{}}10\%\\ 3060 labels\\ 30596 images\end{tabular}} & \multicolumn{2}{c}{\begin{tabular}[c]{@{}c@{}}20\%\\ 6125 labels\\ 30596 images\end{tabular}} \\ \midrule
Supervised (ICCV'19)         & \multicolumn{2}{c}{41.13}                                                                   & \multicolumn{2}{c}{50.32}                                                                   & \multicolumn{2}{c}{53.99}                                                                     & \multicolumn{2}{c}{58.06}                                                                     \\ \midrule
Mean-Teacher (NeurIPS'17)     & \multicolumn{2}{c}{43.36 \rise{2.23}}                                                            & \multicolumn{2}{c}{51.92 \rise{1.60}}                                                             & \multicolumn{2}{c}{55.93 \rise{1.94}}                                                              & \multicolumn{2}{c}{58.88 \rise{0.82}}                                                              \\
MixMatch (NeurIPS'19)         & \multicolumn{2}{c}{41.77 \rise{0.64}}                                                            & \multicolumn{2}{c}{51.23 \rise{0.91}}                                                            & \multicolumn{2}{c}{52.62 \drop{1.37}}                                                              & \multicolumn{2}{c}{57.43 \drop{0.63}}                                                              \\
FixMatch (NeurIPS'20)         & \multicolumn{2}{c}{42.44 \rise{1.31}}                                                            & \multicolumn{2}{c}{51.89 \rise{1.57}}                                                            & \multicolumn{2}{c}{55.79 \rise{1.80}}                                                               & \multicolumn{2}{c}{59.63 \rise{1.57}}                                                              \\ \midrule
SSP3D (ours)                  & \multicolumn{2}{c}{\textbf{46.99 \Rise{5.86}}}                                                   & \multicolumn{2}{c}{\textbf{55.23 \Rise{4.91}}}                                                   & \multicolumn{2}{c}{\textbf{58.98 \Rise{4.99}}}                                                     & \multicolumn{2}{c}{\textbf{61.64 \Rise{3.58}}}                                                     \\ \midrule
\end{tabular}
}
\label{table1}
\end{table}

\noindent\textbf{Results on ShapeNet and Pix3D} As shown in Table~\ref{table1}, we compare our method with the supervised-only models under the settings of 1\%, 5\%, 10\% and 20\% labeled data. The experimental results show that our model outperforms supervised baselines by clear margins, especially under the setting with only 1\% labels where our model outperforms the supervised model by 5.86\%. Notably, our model outperforms the latest SOTA method FixMatch~\cite{sohn2020fixmatch} by 4.55\% with only 1\% labeled data, demonstrating that directly extending existing SSL methods is sub-optimal for the task of single-view 3D reconstruction and that the proposed prototype attentive module and shape naturalness module are effective.

\begin{figure}[ht]
\centering
\includegraphics[width=0.9\columnwidth]{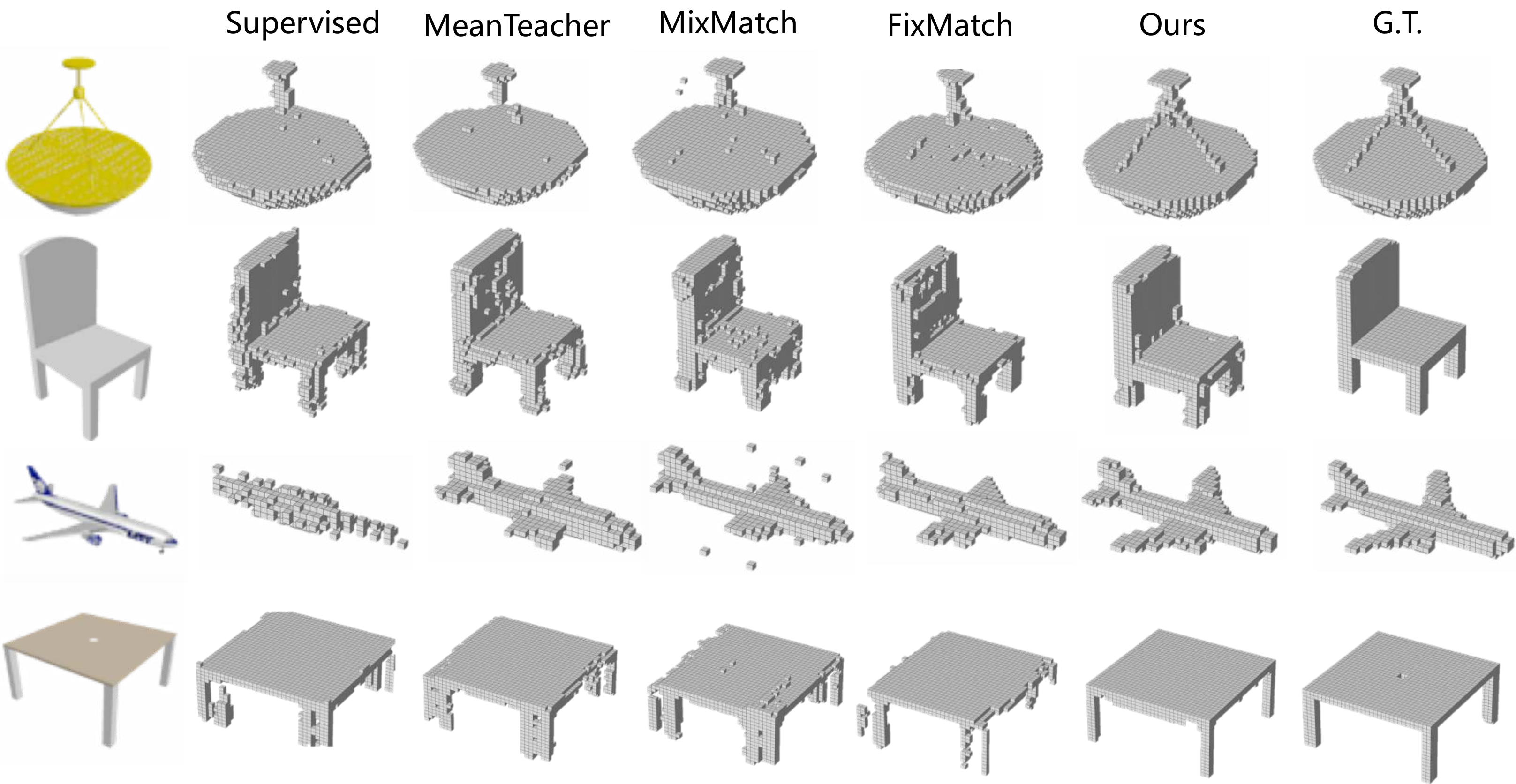} 
\caption{Examples of single-view 3D Reconstruction on ShapeNet with 5\% labels. 
}
\label{fig04}
\end{figure}

\begin{table*}[]
\caption{\textbf{Comparisons of single-view 3D object reconstruction on Pix3D at $128^3$ resolution with 10\% supervised data.} We report the mean IoU $(\%)$ of all categories. The best number for each category is highlighted in bold.}
\centering
\scalebox{0.75}{
\begin{tabular}{ccccccccccc}
\toprule
Approach\textbackslash split            & \begin{tabular}[c]{@{}c@{}}chair\\ 267/2672\end{tabular} & \begin{tabular}[c]{@{}c@{}}bed\\ 78/781 \end{tabular} & \begin{tabular}[c]{@{}c@{}}bookcase\\ 28/282\end{tabular} & \begin{tabular}[c]{@{}c@{}}desk\\ 54/546\end{tabular} & \begin{tabular}[c]{@{}c@{}}misc\\ 4/48\end{tabular} & \begin{tabular}[c]{@{}c@{}}sofa\\ 153/1532\end{tabular} & \begin{tabular}[c]{@{}c@{}}table\\ 145/1451\end{tabular} & \begin{tabular}[c]{@{}c@{}}tool\\ 3/36\end{tabular} & \begin{tabular}[c]{@{}c@{}}wardrobe\\ 18/189\end{tabular} & \textbf{Mean}                                                   \\ \midrule
Supervised~\cite{pix2vox}  & 19.27                                                                  & 32.10                                                               & 23.99                                                                   & 25.32                                                               & 18.37                                                             & 62.29                                                                 & 22.77                                                                  & 11.32                                                             & 81.88                                                                   & 29.80                                                           \\
\midrule
MeanTeacher~\cite{meanteacher} & \begin{tabular}[c]{@{}c@{}}21.66\\ \rise{2.39}\end{tabular}                & \begin{tabular}[c]{@{}c@{}}35.04\\ \rise{2.94}\end{tabular}             & \begin{tabular}[c]{@{}c@{}}18.88\\ \drop{5.11}\end{tabular}                 & \begin{tabular}[c]{@{}c@{}}26.17\\ \rise{0.85}\end{tabular}             & \begin{tabular}[c]{@{}c@{}}22.37\\ \rise{4.00}\end{tabular}           & \begin{tabular}[c]{@{}c@{}}64.19\\ \rise{1.90}\end{tabular}               & \begin{tabular}[c]{@{}c@{}}24.03\\ \rise{1.26}\end{tabular}                & \begin{tabular}[c]{@{}c@{}}9.18\\ \drop{2.14}\end{tabular}            & \begin{tabular}[c]{@{}c@{}}84.34\\ \rise{2.46}\end{tabular}                 & \begin{tabular}[c]{@{}c@{}}31.40\\ \rise{1.60}\end{tabular}         \\
\midrule
FixMatch~\cite{sohn2020fixmatch}    & \begin{tabular}[c]{@{}c@{}}21.95\\  \rise{2.68}\end{tabular}                & \begin{tabular}[c]{@{}c@{}}26.69\\ \drop{5.41}\end{tabular}             & \begin{tabular}[c]{@{}c@{}}16.06\\ \drop{7.93}\end{tabular}                 & \begin{tabular}[c]{@{}c@{}}22.12\\ \drop{3.20}\end{tabular}             & \begin{tabular}[c]{@{}c@{}}17.87\\ \drop{0.50}\end{tabular}           & \begin{tabular}[c]{@{}c@{}}63.74\\ \rise{1.45}\end{tabular}               & \begin{tabular}[c]{@{}c@{}}20.64\\ \drop{2.13}\end{tabular}                & \begin{tabular}[c]{@{}c@{}}6.89\\ \drop{4.43}\end{tabular}            & \begin{tabular}[c]{@{}c@{}}84.45\\ \rise{2.57}\end{tabular}                 & \begin{tabular}[c]{@{}c@{}}30.35\\ \rise{0.55}\end{tabular}         \\
\midrule
SSP3D (ours) & \textbf{\begin{tabular}[c]{@{}c@{}}23.97\\ \Rise{3.04}\end{tabular}}       & \textbf{\begin{tabular}[c]{@{}c@{}}46.33\\ \Rise{13.36}\end{tabular}}   & \textbf{\begin{tabular}[c]{@{}c@{}}32.77\\ \Rise{7.01}\end{tabular}}        & \textbf{\begin{tabular}[c]{@{}c@{}}32.89\\ \Rise{4.76}\end{tabular}}    & \textbf{\begin{tabular}[c]{@{}c@{}}24.35\\ \Rise{2.96}\end{tabular}}  & \textbf{\begin{tabular}[c]{@{}c@{}}68.32\\ \Rise{5.76}\end{tabular}}      & \textbf{\begin{tabular}[c]{@{}c@{}}23.84\\ \Rise{2.13}\end{tabular}}       & \textbf{\begin{tabular}[c]{@{}c@{}}39.06\\ \Rise{27.58}\end{tabular}} & \textbf{\begin{tabular}[c]{@{}c@{}}89.59\\ \Rise{6.88}\end{tabular}}        & \textbf{\begin{tabular}[c]{@{}c@{}}35.39\\ \Rise{5.59}\end{tabular}} \\ \bottomrule
\end{tabular}
\label{table2}
}
\end{table*}

We further conduct experiments on Pix3D~\cite{sun2018pix3d}. The experimental results are shown in Table \ref{table2}. Considering that the 3D voxel resolution of Pix3D is $128^3$, which increases the model complexity, we compare it with the supervised method and the two state-of-the-art methods of MeanTeacher~\cite{meanteacher} and FixMatch~\cite{sohn2020fixmatch} only under the setting of 10\% labeled data due to limited GPU resources. We report the reconstruction performance of each category. For some categories with few labeled data (\emph{e.g.}, tool and bed), our model outperforms supervised models by 27.58\% and 13.36\%. Due to the scarcity of annotated training data and that the other two methods (MeanTeacher and FixMatch) do not have the guidance of strong shape priors, they do not perform as well as supervised methods. Overall, our method outperforms supervised methods by 5.59\% measured by on the mean IoU of all categories. Compared with MeanTeacher~\cite{meanteacher}, it is also better by 4.99\%. 

\begin{figure}[ht]
\centering
\includegraphics[width=0.82\columnwidth]{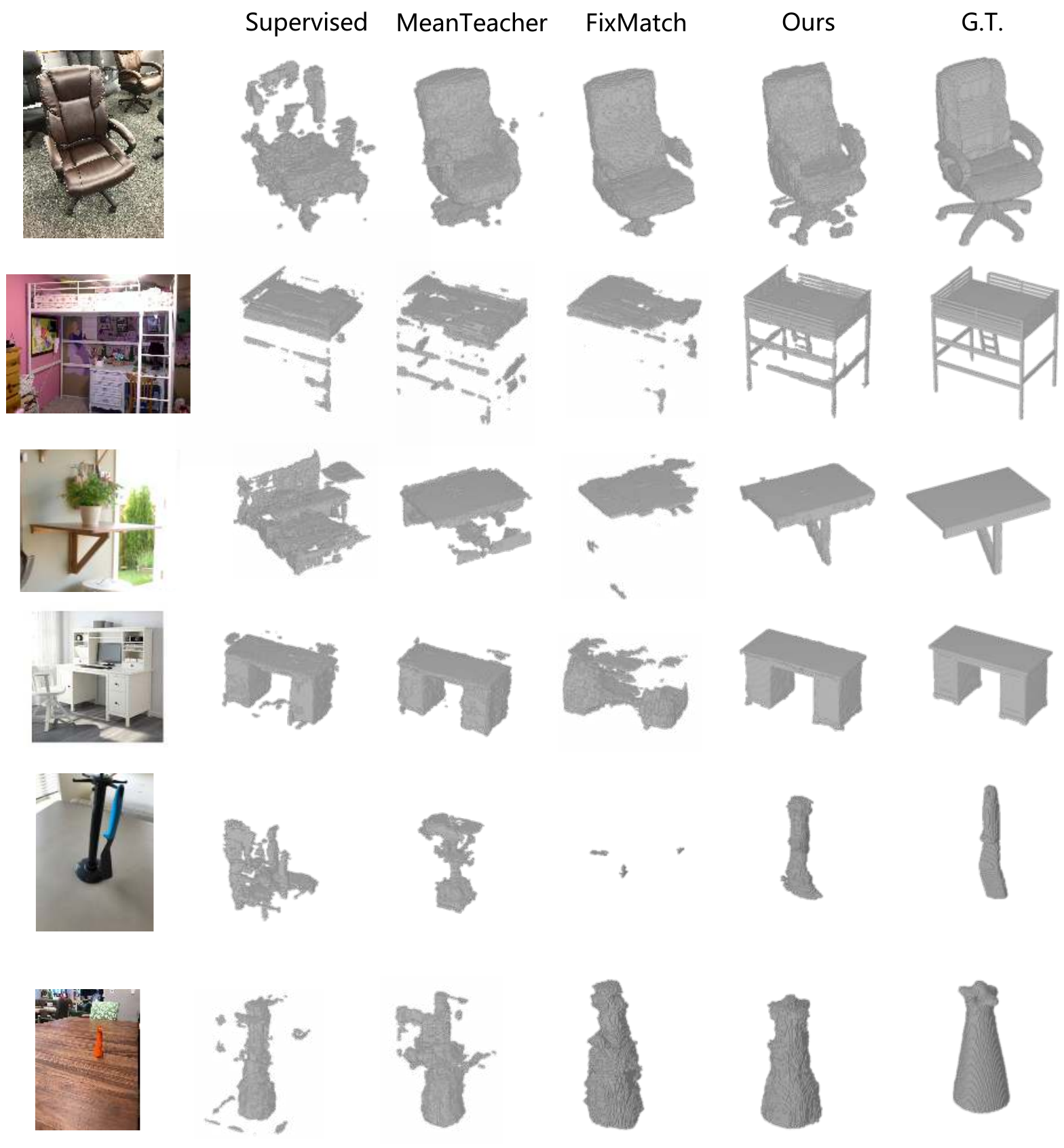} 
\caption{Examples of single-view 3D Reconstruction on Pix3D with 10\% labels. 
}
\label{fig05}
\end{figure}

We further provide qualitative results on both datasets in Fig.~\ref{fig04} and Fig.~\ref{fig05}. As can be seen from Fig.~\ref{fig04}, for images with clean background, our method produces a smoother object surface than baseline methods. For data with complex background and heavy occlusions in Fig.~\ref{fig05}, shapes generated by our method are much better than alternative methods.

\noindent\textbf{Transferring to Novel Category Results}
Wallace \emph{et al.}~\cite{wallace2019few} propose a few-shot setting for single-view 3D reconstruction via shape priors. 
We also train the model with seven base categories and finetune the model with only 10 labeled data in the novel categories. During inference, we report the performance on novel categories. We also compare our method with CGCE~\cite{cgce} and PADMix~\cite{padmix}, as shown in Table~\ref{table3}. Under the 10-shot setting, our method outperforms the zero-shot baseline by 12\%. We hypothesize that the improvement is mainly due to our more reasonable shape prior module design as well as the usage of a large number of unlabeled data. 

\begin{table*}[]
\caption{Comparison of single-view 3D object reconstruction on novel categories of ShapeNet at $32^3$ resolution under 10-shot setting. We report the mean IoU(\%) per category. The best number for each category is highlighted in bold.}
\centering
\scalebox{0.95}{

\begin{tabular}{@{}c|c|c|c|c|c|c|c@{}}
\toprule
          & cabinet         & sofa & bench & watercraft & lamp & firearm & \textbf{Mean} \\ \midrule
Zero-shot & 69            & 52 & 37         & 28            & 19         & 13    & 36 \\ \midrule
Wallace (ICCV'19)~\cite{wallace2019few}   & 69 \rise{0}    & 54 \rise{2} & 36 \drop{1}     & 36 \rise{8}        & 19 \rise{0} & 24 \rise{11} & 39 \rise{3} \\
CGCE (ECCV'20)~\cite{cgce} & 71 \rise{2}  & 54 \rise{2} & 37 \rise{0}     & 41 \rise{13}       & 20 \rise{1} & 23 \rise{10}    & 41 \rise{5} \\ 
PADMiX (AAAI'22)~\cite{padmix} &66 \drop{3} &57 \rise{5} &41 \rise{4} &46 \rise{18} &31 \rise{12} &39 \rise{26} &47 \rise{11} \\
\midrule
SSP3D(ours)      & \textbf{72} \Rise{3}    & \textbf{61} \Rise{9} & \textbf{43} \Rise{6}  & \textbf{49} \Rise{21}       & \textbf{31} \Rise{12} & \textbf{34} \Rise{21}    & \textbf{48} \Rise{12} \\ \bottomrule
\end{tabular}
}
\label{table3}
\end{table*}

\subsection{Ablation Study}
In this section, we evaluate the effectiveness of proposed modules and the impact of hyper-parameters. The experiments are under the 1\% ShapeNet setting and 10\% Pix3D setting if not mentioned elsewhere.

\noindent\textbf{Prototype Attentive Module}
Here we analyze the effectiveness of shape prior module in 3D reconstruction. To do this, we compare our method with various prior aggregation methods including totally removing the prototype attention module (w/o PAM), fusing class-specific prototypes through averaging (w. average) and using LSTM~\cite{lstm} fusion for prototype shape priors.
As shown in Table~\ref{table4}, removing the prototype attentive module results in a large drop of 3.55\% and 4.17\% in performance on both datasets , demonstrating the effectiveness of using class-specific shape priors for single-view 3D reconstruction. Our prior module also outperforms all other prior aggregation methods, indicating that self-attention mechanism is better at capturing the relationships between input images and class-specific prototype shape priors.

\begin{figtab}

  \begin{minipage}[b]{0.65\linewidth}
    \centering
    \tabcaption{Ablation study of different modules and losses. We report the mean IoU(\%) of both datasets.}
    \scalebox{0.61}{
    
    \begin{tabular}{@{}c|cccccccc|c|c@{}}
    \toprule
              & PAM & average & LSTM & SNM & $ {\cal{L}}_{unsup}$ & $ {\cal{L}}_{BCE}$ & $ {\cal{L}}_{rec}$ & score & ShapeNet & Pix3D \\ \midrule
    baseline  &     &         &      &     &    &      & \checkmark    &       & 41.13 \drop{5.86}          & 29.80 \drop{5.59}       \\ \midrule
    w/o PAM   &     &         &      & \checkmark   & \checkmark  &      & \checkmark    & \checkmark     & 43.44 \drop{3.55}          & 31.32 \drop{4.17}       \\
    w average   &     & \checkmark       &      & \checkmark   & \checkmark  &      & \checkmark    & \checkmark     & 43.80 \drop{3.19}   & 32.64 \drop{2.75}  \\
    w LSTM      &     &         & \checkmark    & \checkmark   & \checkmark  &      & \checkmark    & \checkmark     & 44.61 \drop{2.38} & 33.42 \drop{1.97}\\\midrule
    w/o SNM   & \checkmark   &         &      &     & \checkmark  &      & \checkmark    & -     & 45.87 \drop{1.12}         & 33.90 \drop{1.49}    \\
    w/o score & \checkmark   &         &      & \checkmark   & \checkmark  &      & \checkmark    &       & 46.32 \drop{0.67}         & 34.62 \drop{0.77}      \\
    w $ {\cal{L}}_{BCE}$& \checkmark   &         &      & \checkmark   &    & \checkmark    & \checkmark    & \checkmark& 45.85 \drop{1.14}  & 34.02 \drop{1.37}   \\ \midrule
    SSP3D(ours)      & \checkmark   &         &      & \checkmark   & \checkmark  &      & \checkmark    & \checkmark     & \textbf{46.99}          & \textbf{35.39}       \\ \bottomrule
    \end{tabular}
    }\quad \quad \quad 
    \label{table4}
  \end{minipage}
    \begin{minipage}[b]{0.31\linewidth}
    \centering
    \includegraphics[width=1.0\linewidth]{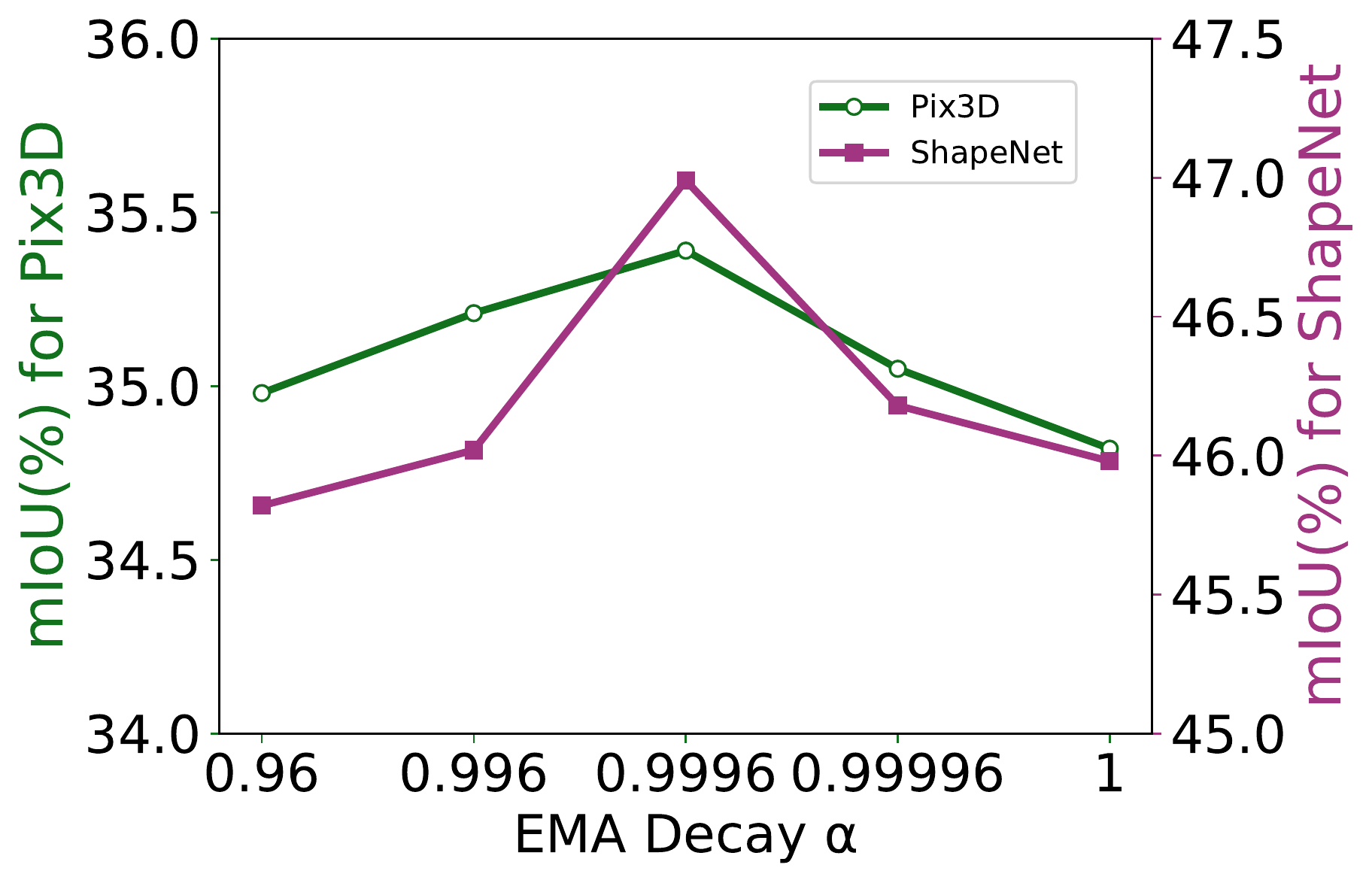} 
    \figcaption{Ablation study of different EMA decay $\alpha$. 
    }
    \label{fig06}
  \end{minipage}
\end{figtab}

\noindent\textbf{Shape Naturalness Module}
In order to demonstrate the effectiveness of the shape naturalness module, we remove the module (w/o SNM), that is, remove the GAN loss ${\cal{L}}_d$   and only use $ {\cal{L}}_{rec}$ to optimize the network in the warm-up stage. In addition, we also verify the effectiveness of the confidence scores generated by the discriminator through replacing all the confidence scores as 1 (w/o score) and check the performance. Experiments show that the performance drops 1.12\% and 0.67\% on ShapeNet without SNM and scorer respectively, indicating that SNM plays an important role in our framework, which may avoid unnatural 3D shape generation, and the confidence score could avoid the negativeness of noisy or biased labels.

\noindent\textbf{EMA and Loss}
We also verify the effect of EMA. In our experiments, we find that EMA decay coefficient $\alpha=0.9996$ gives the best validation performance. As shown in Fig.~\ref{fig06}, the performance slightly drops at different decay rates. For the unsupervised loss function in the Teacher-student mutual learning stage, if the unsupervised squared L2 loss is replaced by binary cross entropy (w $ {\cal{L}}_{BCE}$), the performance of the model will also drop 1.14\% and 1.37\% in performance on two datasets shown in Table~\ref{table4}.

\section{Conclusion}
We introduced SSP3D, which is the first semi-supervised approach for single-view 3D reconstruction. We presented an effective prototype attentive module for semi-supervised setting to cope with limited annotation data. We also used a discriminator to evaluate the quality of pseudo-labels so as to generate better shapes. We conducted extensive experiments on multiple benchmarks and the results demonstrate the the effectiveness of the proposed approach. In future work, we would like to explore the semi-supervised setting on other 3D representation, such as mesh or implicit function.

\noindent\textbf{Acknowledgement} Y.-G. Jiang was
sponsored in part by ``Shuguang Program'' supported by Shanghai Education Development Foundation and Shanghai Municipal
Education Commission (No. 20SG01). Z. Wu was supported by NSFC under Grant No. 62102092.

%
\bibliographystyle{splncs04}
\bibliography{egbib}
\end{document}